\documentclass{article}
\usepackage[eatrack]{log_2024}				%

\usepackage{booktabs}						%
\usepackage{multirow}						%
\usepackage{amsfonts}						%
\usepackage{graphicx}						%
\usepackage{duckuments}						%

\usepackage{mathtools}
\usepackage{pifont}
\usepackage{siunitx}
\usepackage{algorithm}
\usepackage{algpseudocode}

\usepackage[numbers,compress,sort]{natbib}	%

\title[Domain and Range Aware Negatives for KGE]{Domain and Range Aware Synthetic Negatives Generation for Knowledge Graph Embedding Models}

\author[Bernardi, A. and Costabello, L.]{%
Alberto Bernardi\\
Accenture Labs, Dublin\\
\email{alberto.bernardi@accenture.com}\And
Luca Costabello\\
Accenture Labs, Dublin\\
\email{luca.costabello@accenture.com}
}

\begin{document}

\maketitle

\begin{abstract}
Knowledge Graph Embedding models, representing entities and edges in a low-dimensional space, have been extremely successful at solving tasks related to completing and exploring Knowledge Graphs (KGs). One of the key aspects of training most of these models is teaching to discriminate between true statements (\textit{positives}) and false ones (\textit{negatives}). However, the way in which negatives can be defined is not trivial, as facts missing from the KG are not necessarily false and a set of ground truth negatives is hardly ever given. This makes synthetic negative generation a necessity. Different generation strategies can heavily affect the quality of the embeddings, making it a primary aspect to consider. We revamp a strategy that generates corruptions during training respecting the domain and range of relations, we extend its capabilities and we show our methods bring substantial improvement (+10\% MRR) for standard benchmark datasets and over +150\% MRR for a larger ontology-backed dataset.
\end{abstract}

\section{Introduction}

Knowledge Graphs (KGs) are a flexible and scalable data representation paradigm to represent complex relationships between different concepts~\cite{hogan2021survey}. Their applicability to model information related to any domain made them reach a significant size, including millions or even billions of edges. This makes manual curation almost impossible and leaves them incomplete and error-prone~\cite{dong2014knowledge_vault}.

In this context, a machine-based approach is necessary to infer missing links and assess the correctness of given links. Pioneer research efforts~\cite{bordes2013transe,nickel2012factorizing} led to Knowledge Graph Embedding (KGE) models, that learn low-dimensional representations for entities and relations to perform a variety of tasks. %

Although some approaches rely uniquely on the true facts that make up a KG, the dominant training approach for KGE models teaches to discriminate between true statements (\textit{positives}) and false ones (\textit{negatives}). However, statements missing in a KG are not necessarily false, and rarely ground truth negatives are provided. This makes the generation of synthetic negatives, or \textit{Negative Sampling} (NS), necessary and determinant during training. A lot of research was carried out, addressing the difficult challenge of coming up with effective and efficient strategies to sample true negatives~\cite{kamigaito2022survey,madushanka2024survey}.

Our work extends an existing NS strategy~\cite{kwompass2015type_constrained} that leverages schema-based information in order to enhance the quality of the negatives. Despite nailing the possibility of using entity type information to generate negatives that are semantically relevant, they neglect how, in case a class has few instances, the likelihood of resampling the same entity from the class increases, so we might end up generating non-diverse and (if the sampled entity coincides with the ground truth one) false negatives. Combining it with random uniform sampling~\cite{bordes2013transe}, we mitigate this issue while also limiting the risk of generating trivial random negatives. The latter problem, indeed, affects the uniform random sampling applied on its own, especially in KGs where some node types are over-represented.

As an additional contribution, we validate our results on two popular benchmark datasets, where the classes of nodes are inferred based on relation types, and on a biological dataset for which classes are available as part of the ontology. Importantly, we limit our analysis to four traditional KGE models. Indeed, despite recent works have claimed new state-of-the-art results, the performance gain is often marginal and inflated~\cite{ruffinelli2020you_can_teach} by computationally expensive training protocols (\textit{e.g.}, 1-N scoring~\cite{dettmetters2018conve}, high-dimensional embeddings) that are hardly scalable to real-world applications. Therefore, we re-implement and explore models of manageable size, showing how our approach can lead to significant improvements while competing with the state-of-the-art. 

\section{Related Work}
\label{sec:related_work}
\paragraph{Knowledge Graph Embedding Models} KGE models learn continuous representations of entities and relations in the KG from the graph topology and its soft regularities (symmetries and asymmetries, concept homophily, long-range relationships...) to solve a wide variety of tasks. We refer the interested reader to~\cite{cao2024survey_kge} for a comprehensive overview.

\paragraph{Negative Sampling} A lot of methods addressed the challenge of generating negative samples from existing KGs. We can broadly categorize them in 4 groups. (i) Static NS methods consider a pool of possible negatives that does not change during training. Within this group we have: the preeminent random sampling, that corrupts either the tail or the head of a triple picking uniformly at random any entity in the KG~\cite{bordes2013transe}, that might be extended to relations~\cite{xu2018rels_corruption} or limited to entities within the mini-batch~\cite{lerer2019pytorch_bigGraph}; probabilistic negative sampling methods, that fix a probability to decide whether to corrupt the head or the tail of a triple~\cite{wang2014transH,xiw2017interpretable,zhang2019novel_negative,cao2021novel_negative} or to use random negatives or those sampled from a predetermined list~\cite{kanojia2017enhancing}; methods that leverage a pre-trained external models (e.g., a pre-trained KGE~\cite{kotnis2017analysis} or Language Model~\cite{mohtashim2022LLM_negatives}) to generate corruptions that are similar to the entity to corrupt; auxiliary data-based, that leverage type information~\cite{kwompass2015type_constrained}, domain or range information~\cite{wang2022kgboost,li2019incorporating}, or the semantics~\cite{weyns2020conditional} of a KG to generate negatives.
(ii) Self-Adversarial NS strategies use the KGE embeddings to model a probability distribution from which to sample a set of negatives. For example, \cite{sun2019rotate} makes the distribution of negatives proportional to the score assigned from the KGE model to the negatives, so to guarantee that, as training progresses, the model is influenced more by more plausible negatives. However, this strategy comes with the risk of assigning higher relevance to false negatives. Other works along this line are~\cite{shan2018confidence,zhang2019nscaching,lei2019adversarial}.
(iii) Adversarial NS methods leverage an external model, trained in a framework that resembles Generative Adversarial Networks~\cite{goodfellow2014gans}, to generate negatives. (iv) Dynamic negative sampling methods, finally, try to inform the generation of negatives by changing their distribution as training progresses to avoid trivial negatives. However, (iii) and (iv) introduce a significant computational overhead. For non-attributive graphs, \cite{yan2023advances} use a similarity score to sample harder negatives, but its efficient definition in KGs is complicated by their heterogeneity. For more details and negative sampling methods, refer to~\cite{kamigaito2022survey,madushanka2024survey}.

\section{Methods}
\subsection{Background and Notation}
A Knowledge Graph $\mathcal{G}=\{ (s,p,o)\} \subseteq \mathcal{E} \times \mathcal{R} \times  \mathcal{E}$ is a set of triples $t=(s,p,o)$, each including a subject (\textit{head}) $s \in \mathcal{E}$, a predicate $p \in \mathcal{R}$, and an object (\textit{tail}) $o \in \mathcal{E}$, where $\mathcal{E}$ and $\mathcal{R}$ are the sets of all entities and relation types, respectively. We refer to the task of predicting unseen triples in a KG as \textit{Link Prediction}. It is formalised in literature as a learning to rank problem, where the objective is learning a scoring function $f: \mathcal{E} \times \mathcal{R} \times \mathcal{E} \rightarrow \mathbb{R}$
that, given an input triple $t=(s,p,o)$, assigns a score $f(t) = f((s,p,o)) \in \mathbb{R}$ proportional to the likelihood that the fact $t$ is true. Predictions of true triples $t^+ \in \mathcal{T}^+$ are ranked against predictions for synthetic negatives $t^- \in \mathcal{T}^-$, to gauge how well the model discriminates true from false statements.
Finally, we introduce the concept of \textit{class}. If in KGs this concept is formally tied to the definition of a schema or an ontology, we will more broadly refer to it to also indicate the grouping of entities we can infer from the relation types in case the ontology is missing. In these cases, we are going to define two classes for every relation type $p \in \mathcal{R}$: $\texttt{domain}_p \coloneq \left\lbrace e \in \mathcal{E} \mid \exists o \in \mathcal{E} \text{ s.t. } (e, p, o) \in \mathcal{G} \right\rbrace$,
$\texttt{range}_p \coloneq \left\lbrace e \in \mathcal{E} \mid \exists s \in \mathcal{E} \text{ s.t. } (s, p, e) \in \mathcal{G} \right\rbrace$.

\subsection{Ontology-based Negative Sampling}
The dominant approach to NS while training a KGE model is, given a positive triple $t=(s,p,o)$, to corrupt either the subject or the object with $\eta$ entities sampled uniformly at random from $\mathcal{E}$~\cite{bordes2013transe}. Formally, the set of negatives to sample from is defined as $\mathcal{T}_{rand}^- = \left\lbrace t^- = (s', p, o) \mid s' \in \mathcal{E} \right\rbrace \cup \left\lbrace t^- = (s, p, o') \mid o' \in \mathcal{E} \right\rbrace$. Notice that $\mathcal{T}_{rand}^- \cap \mathcal{G} \neq \emptyset$, which implies that the above approach might lead to the generation of false negatives. Despite this might slightly slow down the convergence of the training, it allows an extremely efficient generation. The negatives sampled from $\mathcal{T}_{rand}^-$, however, might be heavily affected by an unbalanced class distribution. Indeed, if an entity type is over-represented in a KG, it will also be over-represented in $\mathcal{T}_{rand}^-$. Therefore, entities outside this class respecting the ``domain'' or ``range'' of a relation might not appear in any of the sampled corruptions, thus significantly limiting the learning process. To address this issue, we take inspiration from~\cite{kwompass2015type_constrained}, that proposes to corrupt triples respecting the semantics of relations extracted from the ontology (if given) or from the facts in the training data. Formally, given triple $t=(s,p,o)$, the set of negatives for this approach will be denoted $\mathcal{T}_{t.c.}^- = \left\lbrace (s', p, o) \mid s' \in \texttt{domain}_p \right\rbrace \cup \left\lbrace (s, p, o') \mid o' \in \texttt{range}_p \right\rbrace$, where \textit{t.c.} stands for Type-Constrained. \newline
Unlike~\cite{kwompass2015type_constrained}, we do not limit the NS to  $\mathcal{T}_{t.c.}^-$; instead, we also sample a proportion of negatives (determined by a hyper-parameter $\nu$) from  $\mathcal{T}_{rand}^-$ (refer to Appendix~\ref{apx:alg_our_NS} for the pseudo-code). Otherwise, in case we had a class including few entities, the corruptions obtained replacing entities within this class would likely be identical and might have led to a significant number of false negatives. In the next sections, we are going to analyse in detail how much this affects KGs with different class distributions and show how our strategy allows to significantly improve results compared to the default approaches.

\section{Experiments}
\subsection{Experimental Settings}
\paragraph{Datasets} We evaluate our NS strategy on three KGs: FB15k-237~\cite{toutanova2015observed} and WN18RR~\cite{dettmetters2018conve} are two of the most widely used benchmarks for link prediction; Hetionet~\cite{himmelstein2017hetionet} is a network encoding biological data. Differently from the former two, Hetionet does not come with pre-defined train-test splits. Therefore, we create a test set of 10,000 triples and a validation set of 5,000 extracted at random\footnote{We are aware that properly splitting Hetionet poses multiple challenges~\cite{bonner2021understanding}. However, the purpose of our analysis is to gauge the relative impact of NS strategies on the performance of the models and not to assess the absolute performance of the models. Therefore, we opted for a simple, random split.} from the training set so that no entity of the test and validation sets was not present in the training set. A summary of the statistics is available in Appendix~\ref{apx:datasets}.

\paragraph{Evaluation}
We evaluate our models in the \textit{filtered} setting: for every target triple $t = (s,p,o)$, we separately corrupt the subject and the object with all the entities in the KG not generating a fact in the training, validation or test sets. We then rank test triples against all corruptions. The metrics used are the usual for link prediction: Mean Reciprocal Rank (MRR) and Hits at N (Hits@N).

\paragraph{Hyperparameter Search} We perform an extensive grid search to validate our results. For FB15k-237 and WN18RR, we explore embedding dimension $k=[\num{200}, \num{350}]$, $\eta = [\num{10}, \num{20}, \num{30}]$, loss functions $[\text{self-adversarial}, \text{multi-class NLL}]$. As optimizer we use Adam~\cite{bengio2015adam} with learning rate $lr=[\num{1e-2}, \num{1e-3}, \num{1e-4}, \num{1e-5}]$ and $\mathcal{L}^p$-regularization coefficient $\lambda = [\num{1e-2}, \num{1e-3}, \num{1e-4}]$. Finally, we test the proportion $\nu$ of negatives sampled respecting domain and range constraints: $\nu = [0.0-1.0]$ with step $0.1$. We run experiments for \num{1000} epochs on FB15k-237 and \num{4000} epochs on WN18RR, enabling early stopping monitoring MRR, with patience set to \num{10} and validation frequency of \num{25} epochs.
For Hetionet we run experiments for $1000$ epochs and explore $k = [\num{100}, \num{200}]$ and $\eta = [\num{10}, \num{20}]$, while we keep the rest of the grid search unchanged.

\paragraph{Baseline}
We compare our NS strategy with KGE models trained following the random uniform NS~\cite{bordes2013transe} and the Type-Constrained approach~\cite{kwompass2015type_constrained} we have extended. For a fair comparison, we pick as the best baselines those that share the same (or smaller) embedding size and number of negatives of the best performing model trained with our own NS strategy. The KGE models we consider are TransE~\cite{bordes2013transe}, DistMult~\cite{yang2015distmult}, ComplEx-N3~\cite{trouillon2016complex,lacroix2018complex_n3} and RotatE~\cite{sun2019rotate}. Despite some recent works (among the others~\cite{balazevic2019tucker,zhu2022scalable}) claim new state-of-the-art results, the gain is often marginal and inflated by expensive training protocols. Moreover, the impact of our strategy can be gauged independently on the models considered. We leave for future work the analysis of a wider set of models.

\subsection{Main Results}
The results of our experiments are reported in Table~\ref{tab:results}. Our NS method outperforms the random approach and the Type-Constrained one in all cases, with the only exceptions of TransE and DistMult on WN18RR. The improvement is substantial in FB15k-237: +10\% MRR on the random baseline for ComplEx-N3 and +33\% on the type-constrained baseline. Indeed, some of the classes inferred from relation types for FB15k-237 have extremely reduced cardinality (11 classes have just one node, 24 up to 3 nodes, 56 up to 10), thus making the generation of negatives entirely based on type-constraints harmful, as it leads to many false negatives. On the other hand, on Hetionet, our NS strategy leads to impressive gains of well over +150\% MRR relative to the random sampling baseline. The definition of classes based on a proper ontology, indeed, allows for the generation of more significant and diverse negatives, not dominated by over-represented classes (\textit{e.g.}, Gene and Biological Process). At the same time, the proportion of random uniform negative samples helps increasing the diversity for classes that have few instances (\textit{e.g.}, Disease and Pharmacological Class), thus motivating the improvement over the T.C. baseline.\\
As a final remark, the optimal value of $\eta$ for TransE, DistMult and ComplEx-N3 on WN18RR is \num{30}, while for all the other models is \num{10}. The low (or null) improvement in the performance for the former three models suggests that our strategy impacts more models trained with few negative samples. Indeed, if the number grows, too many domain and range-based negatives can cause issues similar to those outlined for Type-Constrained NS. Random uniform sampling on its own, on the other hand, is enough to yield more meaningful negatives if $\eta$ grows significantly. This shows how our approach can hugely benefit efficiency, allowing for a reduced number of negative samples. Moreover, the overhead introduced by our strategy is minimal, as ComplEx-N3 run on Hetionet (over 2 million triples) took an average ${\sim}5$ms per step extra overhead compared to the random uniform sampling. \\
For the detailed list of the optimal hyperparameters refer to Appendix~\ref{apx:hyperparams}, and for further exploration of the impact of $\nu$ on the results refer to Appendix~\ref{apx:results_vs_nu}.

\begin{table}[t]
    \centering
    \caption{Results for the different Negative Sampling (NS) strategies: ``Rand.'' indicates random uniform sampling, ``T.C.'' the Type-Constrained one. All reported results are obtained with AmpliGraph~\cite{ampligraph}. Random uniform sampling results do not always match SOTA as we deliberately used smaller embedding size and number of negatives for a fair comparison. Best results for each datasets are reported in \textbf{bold}, while across models are \underline{underlined}.}
    \scriptsize
      \begin{tabular}{p{1.2cm}p{.7cm}p{.5cm}p{.4cm}p{.4cm}p{.7cm}p{.5cm}p{.4cm}p{.4cm}p{.7cm}p{.5cm}p{.4cm}p{.4cm}p{.7cm}p{.5cm}}%
        \toprule
         \multirow{2}{*}{\textbf{Model}} & \multirow{2}{*}{\textbf{NS}} & \multicolumn{4}{c}{\textbf{WN18RR}} & \multicolumn{4}{c}{\textbf{FB15k-237}} & \multicolumn{4}{c}{\textbf{Hetionet}} \\
         \cmidrule(lr){3-6} \cmidrule(lr){7-10} \cmidrule(lr){11-14}
         & & \textbf{MRR} & \textbf{H@1} & \textbf{H@3} & \textbf{H@10} & \textbf{MRR} & \textbf{H@1} & \textbf{H@3} & \textbf{H@10} & \textbf{MRR} & \textbf{H@1} & \textbf{H@3} & \textbf{H@10} \\
        \cmidrule(lr){1-1} \cmidrule(lr){2-2} \cmidrule(lr){3-6} \cmidrule(lr){7-10} \cmidrule(lr){11-14}
         \multirow{3}{*}{\textbf{TransE}} & Rand. & \underline{0.25} & \underline{0.04} & \underline{0.42} & \underline{0.56} & 0.29 & 0.20 & 0.33 & 0.47 & 0.04	& 0.02	& 0.04	& 0.08\\
         & T.C. & 0.19 & 0.01 & 0.32 & 0.47 & 0.15 & 0.12 & 0.17 & 0.22 & 0.05 & 0.03 & 0.05 & 0.08 \\
         & Ours & 0.24 & \underline{0.04} & 0.41 & 0.55 & \underline{0.31} & \textbf{0.23} & \textbf{0.35} & \underline{0.49} & \underline{0.23}	& \underline{0.15}	& \underline{0.25}	& \underline{0.38} \\
         \cmidrule(lr){3-6} \cmidrule(lr){7-10} \cmidrule(lr){11-14}
          \multirow{3}{*}{\textbf{DistMult}} & Rand. & \underline{0.47} &	\underline{0.44} &	0.48 &	0.53 & 0.28	& 0.18	& 0.31	& 0.46 & 0.08 & 0.05 & 0.08 & 0.14 \\
         & T.C. & 0.46 &	0.42 &	\underline{0.49} &	\underline{0.55} & 0.22	& 0.16	& 0.24	& 0.35 & \underline{0.22}	& \underline{0.14}	& \underline{0.24}	& \underline{0.36}\\
         & Ours & 0.46 &	0.43 &	0.48 &	0.53 & \underline{0.30}  & \underline{0.22}	& \underline{0.33}	& \underline{0.47} & \underline{0.22}	& \underline{0.14}	& \underline{0.24}	& \underline{0.36} \\
         \cmidrule(lr){3-6} \cmidrule(lr){7-10} \cmidrule(lr){11-14}
          \multirow{3}{*}{\textbf{ComplEx-N3}} & Rand. & 0.50	& 0.46	& 0.52	& 0.57 & 0.29	& 0.19	& 0.32	& 0.48 & 0.09 & 0.05 & 0.16 & 0.33 \\
         & T.C. & \underline{0.51}	& 0.47	& \textbf{0.53}	& \underline{0.58} & 0.24	& 0.17	& 0.26	& 0.38 & 0.24	& 0.16	& \underline{0.27}	& 0.40 \\
         & Ours & \underline{0.51}	& \textbf{0.48}	& \textbf{0.53}	& \underline{0.58} & \textbf{0.32}	& \textbf{0.23}	& \textbf{0.35}	& \underline{0.49} & \underline{0.25}	& \underline{0.17}	& \underline{0.27}	& \underline{0.41} \\
         \cmidrule(lr){3-6} \cmidrule(lr){7-10} \cmidrule(lr){11-14}
          \multirow{3}{*}{\textbf{RotatE}} & Rand. & 0.51	& 0.47	& \textbf{0.53}	& \textbf{0.60} & 0.30       & 0.21       & 0.33       & 0.48 & 0.10	& 0.06	& 0.11	& 0.21\\
         & T.C. & 0.51	& 0.47	& \textbf{0.53}	& \textbf{0.60} & 0.20       & 0.13       & 0.20       & 0.34 & 0.11	& 0.07	& 0.11	& 0.17\\
         & Ours & \textbf{0.52}	& \textbf{0.48}	& \textbf{0.53}	& \textbf{0.60} & \underline{0.31}	& \underline{0.22}	& \textbf{0.35}	& \textbf{0.50} & \textbf{0.26}	& \textbf{0.18}	& \textbf{0.29}	& \textbf{0.42} \\
         \bottomrule
    \end{tabular}
    \label{tab:results}
\end{table}

\section{Conclusion and Future Work}
With this work, we have proposed a simple and effective NS strategy, that can significantly enhance the performance of KGE models and help improve the size, and thus the efficiency, of these models. We have validated results on two popular benchmark KGs and on a biological graph that, given an ontology, was affected the most by our method. The results obtained are extremely promising and, given the flexibility of the approach, we leave for further work its application to a broader set of KGE methods. In addition, we plan to extend the experiments to more datasets, to specifically gauge the impact of ontology-derived classes over relation-inferred ones.

\bibliographystyle{unsrtnat}
\bibliography{reference}
\clearpage

\appendix
\section{Appendix}
\label{apx:alg_our_NS}
\subsection{Ontology-based Negative Sampling Pseudo-Code}
We provide the pseudo-code of our Negative Sampling strategy in Algorithm~\ref{alg:our_NS}.

\begin{algorithm}
\caption{Ontology-based Negative Sampling}\label{alg:our_NS}
\begin{algorithmic}
\Require $t = (s, p, o)$, $\mathcal{T}_{rand}^-$, $\mathcal{T}_{t.c.}^-$, $\nu \in [0, 1]$, $\eta \in \mathbb{N}$.
\Ensure $\mathcal{S}$ set of negatives for $t$ such that $|\mathcal{S}| = \eta$.
\State $\eta_{rand} = \lfloor \eta \cdot \nu \rfloor$ \Comment{Number of negatives to sample uniformly at random}
\State $\eta_{t.c.} = \lceil \eta \cdot (1 - \nu) \rceil$ \Comment{Number of Ontology-based negatives to sample}
\State $\mathcal{S}_{rand} \gets \text{sample}\left(\mathcal{T}_{rand}^-, \eta_{rand} \right)$
\State $\mathcal{S}_{t.c.} \gets \text{sample}\left(\mathcal{T}_{t.c.}^-, \eta_{rand} \right)$ \\
\Return $\mathcal{S} \gets \mathcal{S}_{rand} \cup \mathcal{S}_{t.c.}$
\end{algorithmic}
\end{algorithm}

\subsection{Dataset Statistics}
\label{apx:datasets}
Statistics of the three datasets are reported in Table~\ref{tab:dataset_stats}.

\begin{table}[h]
    \centering
    \caption{Statistics of the datasets.}
    \scriptsize
    \begin{tabular}{lccccccc}
        \toprule
         \textbf{Dataset} & \textbf{\#Entities} & \textbf{\#Relations} & \textbf{\#Train} & \textbf{\#Valid} & \textbf{\#Test} & \textbf{Ontology} & \textbf{\#Classes}  \\
         \midrule
         FB15k-237 & \num{15541} & \num{237} & \num{272115} & \num{17535} & \num{20466} & \ding{55} & \num{474}\\
         WN18RR & \num{40943} & \num{18} & \num{86835} & \num{3034} & \num{3134} & \ding{55} & \num{36} \\
         Hetionet & \num{45158} & \num{24} & \num{2235197} & \num{5000} & \num{10000} & \ding{51} & \num{11} \\
         \bottomrule
    \end{tabular}
    \label{tab:dataset_stats}
\end{table}

\subsection{Implementation Details}
All experiments are implemented with AmpliGraph library~\cite{ampligraph}, using Tensorflow 2.9.0 and Python 3.8.10 in the backend. All our code was run under Ubuntu 20.04 on an Intel Xeon Gold 6226R, 256GB, equipped with Tesla A100 40GB GPUs. 

\subsection{Optimal Hyperparameters}
\label{apx:hyperparams}

In Tables~\ref{tab:hyperparams_wn18rr}-\ref{tab:hyperparams_fb15k237}-\ref{tab:hyperparams_hetionet} we report the best hyperparameters obtained grid searching our models. $k$ specifies the dimension of the KGE embeddings; $\eta$ the number of negatives generated for each training triple; $lr$ is the learning rate of the Adam optimizer used in our experiments; $\nu$ specifies the proportion of the $\eta$ negatives that are sampled respecting the domain and range constraints.

\subsection{Impact of \texorpdfstring{$\nu$}{TEXT} on the Results}
\label{apx:results_vs_nu}

\begin{figure}[t]
    \centering
    \includegraphics[width=0.99\linewidth]{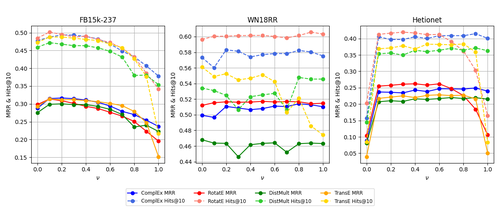}
    \caption{Performance of the different models varying the proportion of ontology-based negatives $\nu$.}
    \label{fig:nu_vs_results}
\end{figure}

As an additional experiment, we report in Figure~\ref{fig:nu_vs_results} the performance of the best models for the different scoring functions, modifying the value of $\nu$, the hyper-parameter specifying the proportion of ontology-based negatives to sample for each training triple.

At a first glance, we can notice how, for the same dataset, the impact of ontology-based sampling has a quite consistent behaviour across all the four models, hinting at its applicability to any other scoring function. Moreover, it looks like there is consistency in the behaviour of the same model across neighbouring values of $\nu$. FB15k-237, for example, favours low values of $\nu$, while Hetionet and WN18RR favour higher values. Importantly, this shows how $\nu$ is an hyper-parameter easy to tune. Indeed, it will be enough to test a couple of extreme values and a central one to have a good estimate of what the impact of the strategy could be.

Another important observation is the difference even a very small $\nu$ can make on the random baseline when there are entity types that are over-represented. It is the case of Hetionet. For this dataset, indeed, we can notice how, even the slightest increase of $\nu$ makes the the performance spike. Indeed, as already mentioned in the main text in the paper, ontology-based negatives allows the generation of negatives more pertinent for each triple and not dominated by entities of the over-represented types. In FB15k-237 and WN18RR the improvement is less marked as there is no class significantly over-represented.

On the other side of the spectrum of values for $\nu$, we can observe that some models see a significant decrease of the performance as we approach the Type-Constrained baseline ($\nu = 1$). This is definitely dataset-dependent, as it is really evident in FB15k-237 and Hetionet where the classes with very few instances are a significant share of the total. However, it looks like the different scoring function impacts this behaviour. Indeed, TransE and RotatE seems to be more badly affected by an higher number of negatives compared to DistMult and ComplEx. As a mere speculation, we conjecture that this might be related to the different nature of the scoring function: translational/rotational in the former cases, and projective in the latter. We propose to better inspect this behaviour in future work, extending the datasets considered in the analysis and the set of models. 

As a final remark, we highlight the inconsistent behaviour of TransE on WN18RR with respect to the other models. The decrease of the Hits@10 observed for higher values of $\nu$ is due to the poor expressive power of the model, that thus struggles at making sense of the harder ontology-based negatives. The poor MRR of the model on this dataset is a further element in support (notice we have not reported the MRR of TransE on WN18RR in Figure~\ref{fig:nu_vs_results} as it was in the range of $[0.03, 0.04]$ for all $\nu$s, including the baseline. Being way below all the other values, it would have compromised the clarity of the plot. However, you can refer to the values reported in Table~\ref{tab:results}).

\clearpage

\begin{table}[htb]
    \centering
    \caption{WN18RR - Best Model Hyperparameters.}
    \scriptsize
    \begin{tabular}{llccccccccc}
         \toprule
         \multirow{2}{*}{\textbf{Model}} & \multirow{2}{*}{\textbf{NS}} & \multicolumn{7}{c}{\textbf{WN18RR}} \\
         \cmidrule(lr){3-10}
         & & $k$ & $\eta$ & Loss & $lr$ & Batch Size & $\mathcal{L}^p$-reg ($p$, $\lambda$) & $\nu$ & Epochs \\
         \midrule
         \multirow{3}{*}{\textbf{TransE}}
         & Rand. & 350 & 30 & Multi-class NLL & \num{1e-4} & \num{1000} & $p=2, \lambda=\num{1e-3}$ & 0.0 & \num{1625} \\
         & T.C. & 350 & 30 & Multi-class NLL & \num{1e-4} & \num{1000} & $p=2, \lambda=\num{1e-3}$ & 1.0 & \num{1625} \\
         & Ours & 350 & 30 & Multi-class NLL & \num{1e-4} & \num{1000} & $p=2, \lambda=\num{1e-3}$ & 0.2 & \num{1500} \\
         \cmidrule(lr){3-10}
         \multirow{3}{*}{\textbf{DistMult}}
         & Rand. & 350 & 30 & Multi-class NLL & \num{1e-4} & \num{5000} & $p=2, \lambda=\num{1e-3}$ & 0.0 & \num{950} \\
         & T.C. & 350 & 30 & Multi-class NLL & \num{1e-4} & \num{5000} & $p=2, \lambda=\num{1e-3}$ & 1.0 & \num{525} \\
         & Ours & 350 & 30 & Multi-class NLL & \num{1e-4} & \num{5000} & $p=2, \lambda=\num{1e-3}$ & 0.6 & \num{650} \\
         \cmidrule(lr){3-10}
         \multirow{3}{*}{\textbf{ComplEx-N3}}
         & Rand. & 200 & 20 & Multi-class NLL & \num{1e-4} & \num{5000} & $p=3, \lambda=\num{1e-2}$ & 0.0 & \num{600} \\
         & T.C. & 200 & 20 & Multi-class NLL & \num{1e-4} & \num{1000} & $p=3, \lambda=\num{1e-2}$ & 1.0 & \num{975} \\
         & Ours & 200 & 20 & Multi-class NLL & \num{1e-4} & \num{1000} & $p=3, \lambda=\num{1e-2}$ & 0.8 & \num{1500} \\
         \cmidrule(lr){3-10}
         \multirow{3}{*}{\textbf{RotatE}}
         & Rand. & 350 & 10 & Self-Adversarial & \num{1e-4} & \num{100000} & $p=3, \lambda=\num{1e-5}$ & 0.0 & \num{1750} \\
         & T.C. & 350 & 10 & Self-Adversarial & \num{1e-4} & \num{50000} & $p=3, \lambda=\num{1e-5}$ & 1.0 & \num{1225} \\
         & Ours & 350 & 10 & Self-Adversarial & \num{1e-4} & \num{100000} & $p=3, \lambda=\num{1e-5}$ & 0.7 & \num{1350} \\
         \bottomrule
    \end{tabular}
    \label{tab:hyperparams_wn18rr}
\end{table}

\begin{table}[htb]
    \centering
    \caption{FB15k-237 - Best Model Hyperparameters.}
    \scriptsize
    \begin{tabular}{llccccccccc}
         \toprule
         \multirow{2}{*}{\textbf{Model}} & \multirow{2}{*}{\textbf{NS}} & \multicolumn{7}{c}{\textbf{FB15k-237}} \\
         \cmidrule(lr){3-10}
         & & $k$ & $\eta$ & Loss & $lr$ & Batch Size & $\mathcal{L}^p$-reg ($p$, $\lambda$) & $\nu$ & Epochs \\
         \midrule
         \multirow{3}{*}{\textbf{TransE}}
         & Rand. & 200 & 10 & Multi-class NLL & \num{1e-3} & \num{100000} & $p=2, \lambda=\num{1e-5}$ & 0.0 & \num{250} \\
         & T.C. & 200 & 10 & Multi-class NLL & \num{1e-3} & \num{100000} & $p=2, \lambda=\num{1e-5}$ & 1.0 & \num{200} \\
         & Ours & 200 & 10 & Multi-class NLL & \num{1e-3} & \num{100000} & $p=2, \lambda=\num{1e-5}$ & 0.1 & \num{1000} \\
         \cmidrule(lr){3-10}
         \multirow{3}{*}{\textbf{DistMult}}
         & Rand. & 350 & 10 & Multi-class NLL & \num{1e-3} & \num{100000} & $p=3, \lambda=\num{1e-3}$ & 0.0 & \num{475} \\
         & T.C. & 350 & 10 & Multi-class NLL & \num{1e-4} & \num{100000} & $p=3, \lambda=\num{1e-3}$ & 1.0 & \num{625} \\
         & Ours & 350 & 10 & Multi-class NLL & \num{1e-3} & \num{100000} & $p=3, \lambda=\num{1e-3}$ & 0.2 & \num{150} \\
         \cmidrule(lr){3-10}
         \multirow{3}{*}{\textbf{ComplEx-N3}}
         & Rand. & 350 & 10 & Multi-class NLL & \num{1e-3} & \num{100000} & $p=3, \lambda=\num{1e-5}$ & 0.0 & \num{275} \\
         & T.C. & 350 & 10 & Multi-class NLL & \num{1e-4} & \num{100000} & $p=3, \lambda=\num{1e-5}$ & 1.0 & \num{550} \\
         & Ours & 350 & 10 & Multi-class NLL & \num{1e-4} & \num{100000} & $p=3, \lambda=\num{1e-3}$ & 0.2 & \num{1000} \\
         \cmidrule(lr){3-10}
         \multirow{3}{*}{\textbf{RotatE}}
         & Rand. & 350 & 10 & Self-Adversarial & \num{1e-4} & \num{100000} & $p=3, \lambda=\num{1e-3}$ & 0.0 & \num{700} \\
         & T.C. & 350 & 10 & Self-Adversarial & \num{1e-4} & \num{100000} & $p=3, \lambda=\num{1e-3}$ & 1.0 & \num{975} \\
         & Ours & 350 & 10 & Self-Adversarial & \num{1e-4} & \num{100000} & $p=3, \lambda=\num{1e-3}$ & 0.1 & \num{500} \\
         \bottomrule
    \end{tabular}

    \label{tab:hyperparams_fb15k237}
\end{table}

\begin{table}[htb]
    \centering
    \caption{Hetionet - Best Model Hyperparameters.}
    \scriptsize
    \begin{tabular}{llccccccccc}
         \toprule
         \multirow{2}{*}{\textbf{Model}} & \multirow{2}{*}{\textbf{NS}} & \multicolumn{7}{c}{\textbf{Hetionet}} \\
         \cmidrule(lr){3-10}
         & & $k$ & $\eta$ & Loss & $lr$ & Batch Size & $\mathcal{L}^p$-reg ($p$, $\lambda$) & $\nu$ & Epochs \\
         \midrule
         \multirow{3}{*}{\textbf{TransE}}
         & Rand. & 200 & 10 & Multi-class NLL & \num{1e-2} & \num{100000} & $p=2, \lambda=\num{1e-5}$ & 0.0 & \num{975} \\
         & T.C. & 200 & 10 & Multi-class NLL & \num{1e-4} & \num{100000} & $p=2, \lambda=\num{1e-3}$ & 1.0 & \num{100} \\
         & Ours & 200 & 10 & Multi-class NLL & \num{1e-4} & \num{100000} & $p=2, \lambda=\num{1e-5}$ & 0.8 & \num{850} \\
         \cmidrule(lr){3-10}
         \multirow{3}{*}{\textbf{DistMult}}
         & Rand. & 200 & 10 & Multi-class NLL & \num{1e-4} & \num{100000} & $p=3, \lambda=\num{1e-5}$ & 0.0 & \num{50} \\
         & T.C. & 200 & 10 & Multi-class NLL & \num{1e-4} & \num{100000} & $p=3, \lambda=\num{1e-3}$ & 1.0 & \num{875} \\
         & Ours & 200 & 10 & Multi-class NLL & \num{1e-4} & \num{100000} & $p=3, \lambda=\num{1e-5}$ & 0.8 & \num{925} \\
         \cmidrule(lr){3-10}
         \multirow{3}{*}{\textbf{ComplEx-N3}}
         & Rand. & 200 & 10 & Multi-class NLL & \num{1e-4} & \num{100000} & $p=3, \lambda=\num{1e-5}$ & 0.0 & \num{25} \\
         & T.C. & 200 & 10 & Multi-class NLL & \num{1e-4} & \num{100000} & $p=3, \lambda=\num{1e-3}$ & 1.0 & \num{750} \\
         & Ours & 200 & 10 & Multi-class NLL & \num{1e-4} & \num{100000} & $p=3, \lambda=\num{1e-3}$ & 0.8 & \num{925} \\
         \cmidrule(lr){3-10}
         \multirow{3}{*}{\textbf{RotatE}}
         & Rand. & 200 & 10 & Multi-class NLL & \num{1e-3} & \num{100000} & $p=3, \lambda=\num{1e-5}$ & 0.0 & \num{25} \\
         & T.C. & 200 & 10 & Multi-class NLL & \num{1e-4} & \num{100000} & $p=3, \lambda=\num{1e-3}$ & 1.0 & \num{375} \\
         & Ours & 200 & 10 & Multi-class NLL & \num{1e-4} & \num{100000} & $p=3, \lambda=\num{1e-5}$ & 0.3 & \num{650} \\
         \bottomrule
    \end{tabular}
    \label{tab:hyperparams_hetionet}
\end{table}

\end{document}